%% file: arxiv.tex
\documentclass[10pt,twocolumn,letterpaper]{article}

\usepackage[pagenumbers]{cvpr} % To produce the REVIEW version
\input{preamble}

\definecolor{cvprblue}{rgb}{0.21,0.49,0.74}
\usepackage[pagebackref,breaklinks,colorlinks,allcolors=cvprblue]{hyperref}

\def\confName{CVPR}
\def\confYear{2027}

\title{RAGA: Refreshable Animatable Gaussian Avatars from a Single Photo}

\author{
  Sadia Mubashshira \quad John Quarles \quad Kevin Desai\\
  {\small\texttt{\{Sadia.Mubashshira, John.Quarles, kevin.desai\}@utsa.edu}}
}

\begin{document}
\raggedbottom
% Existing teaser asset; standard full-width title block, no spacing overrides.
% Append the teaser to CVPR's title block; retain the standard maketitle path.
% Avoid nesting/redefining twocolumn inside its optional argument.
\makeatletter
\apptocmd{\@maketitle}{%
\begin{center}
    \includegraphics[width=\textwidth]{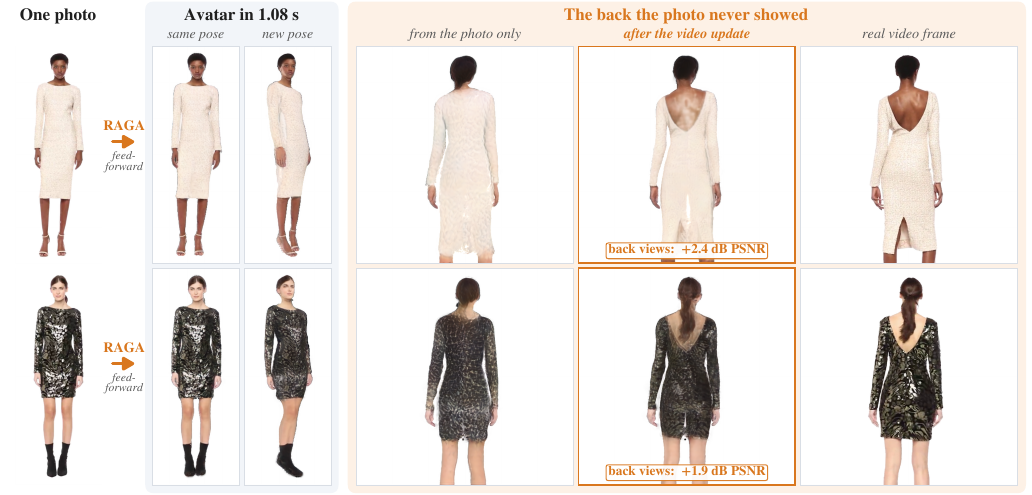}
    \captionsetup{hypcap=false}
    \vspace{-5mm}
    \captionof{figure}{\textbf{Refreshable single-photo avatar.} RAGA creates an animatable full-body Gaussian avatar from one photograph and refreshes it from later images of the same person, without retraining or gradient steps. Surfaces the photo never showed, such as the back of a garment, are recovered from the new observations. The two clips are from UBC-Fashion~\cite{ubcfashion}. Benchmark averages are in \cref{tab:main,tab:farview}.}
    \label{fig:teaser}
\end{center}
}{}{\PackageError{RAGA}{Could not attach the teaser to the title block}{Check that the CVPR article template is loaded.}}
\makeatother
\maketitle

\input{sec/0_abstract}    
\input{sec/1_intro}
\input{sec/2_related_work}
\input{sec/3_method}

\input{sec/4_result}

\input{sec/5_conclusion}
\input{sec/llm_usage}
{
    \small
    \bibliographystyle{ieeenat_fullname}
    \bibliography{main}
}

\clearpage
\appendix

\end{document}

%% file: preamble.tex
\usepackage{graphicx}
\usepackage{cuted}
\usepackage{caption}
\usepackage{multirow}
\usepackage{xcolor}
\usepackage{colortbl}
\usepackage{booktabs}
\makeatletter
\renewcommand{\paragraph}[1]{\par\noindent\textbf{#1}\hspace{0.5em}}
\makeatother

%% file: sec/0_abstract.tex
\begin{abstract}
Avatars created from a single photo are fixed once made: regions the photo never showed, such as the back of a garment, are hallucinated and remain that way. We present Refreshable Animatable Gaussian Avatars (RAGA), a full-body Gaussian avatar that is created from a single image in about one second and can be refreshed whenever new images of the same person become available, without retraining or gradient-based optimization. RAGA maps each new observation into canonical UV space, accumulates it as visibility- and angle-weighted running sums, and re-decodes the fused map with the same decoder used for the photo. Although this decoder is trained only on single images, it makes use of the fused evidence. On 163 clips from five datasets, RAGA created from one photo outperforms LHM, LHM++ and IDOL, including on three datasets never used in training. Refreshing from later frames adds 0.49\,dB PSNR overall and 0.68\,dB on views turned at least $135^\circ$ from the photo, where SSIM and LPIPS also improve. Given the same 16 images on all clips, LHM++ gains 0.15\,dB, against 0.48\,dB for RAGA. Creating an avatar takes 1.08\,s and animating it 48.1\,ms per frame on a laptop RTX~3080 GPU, with body fitting and matting included in creation and pose estimation included in animation. Code will be released on GitHub.
\end{abstract}

%% file: sec/1_intro.tex
\section{Introduction}
\label{sec:introduction}

Avatars that can be created from a single photograph make personalized telepresence accessible: no capture rig, no video and no per-person training~\cite{idol,lhm,guava,lhmpp,forwardrobe}. They share a limitation that follows from the input itself. Large parts of the body and clothing are not visible in one photo, so these methods infer them from learned priors, and once the avatar is created those guesses are final. The back of a dress or the pattern on a sleeve stays hallucinated no matter how much of the person is seen afterwards.

In most uses, however, more of the person does become visible. During a video call or a capture session people turn, gesture and move, revealing surfaces the photo missed. Existing methods can use such evidence in three ways, each with a cost. Multi-image feed-forward methods~\cite{lhmpp,fresa,giga,chatterjee2026realtime} reconstruct from a set of images, but need the images together at creation time and reconstruct again when new ones arrive. Per-subject methods~\cite{gart,gaussianavatar,exavatar,hugs,qian2024gs} optimize an avatar on video, which takes minutes to hours. Feed-forward methods that improve as frames are added exist for head avatars~\cite{fastavatar,ffavatar} and, for full bodies, for re-rendering a performer in the motion observed in a monocular stream~\cite{genfusion}. To our knowledge, none refreshes a full-body avatar that can be driven by new motion.

Practical use also depends on the whole pipeline. Reported reconstruction and animation times usually assume that body pose, camera and foreground mattes are given. Under that assumption, GUAVA~\cite{guava} reconstructs an upper-body avatar in under a second and animates it in real time. With its own per-frame tracker on a laptop GPU, animation takes 7.28\,s per frame in our measurements (\cref{sec:efficiency}). We therefore report timings that include body fitting, matting and per-frame pose estimation.

We present RAGA, a full-body Gaussian avatar that is created from one photo and refreshed as later observations arrive. Our key observation is that a UV-based feed-forward avatar decoder consumes a canonical observation map: image colours and features projected onto the body surface. Later images of the same person can be projected into the same map. RAGA accumulates them as running sums weighted by visibility and viewing angle, a standard fusion rule~\cite{curless1996volumetric,kinectfusion,genfusion}, and re-decodes the fused map with the unchanged decoder. The decoder is trained only on single images, yet it uses the fused evidence, both to fill regions the photo never showed and to revise regions it did. Because only the sums are stored, earlier images can be discarded and a refresh can happen at any time, with the same result as fusing all frames at once.

The refresh mechanism requires neither a new reconstruction backbone nor multi-image training. We implement it on the GUAVA architecture~\cite{guava}, extended from the upper body to the full body with native SMPL-X handling, per-image perspective projection, and full-body training. On 163 clips from five datasets, three of which are never used in training, RAGA created from one photo outperforms the full-body feed-forward methods LHM, LHM++ and IDOL~\cite{lhm,lhmpp,idol}. Refreshing raises PSNR and SSIM on every dataset, and the gains concentrate where they should: on views turned at least $135^\circ$ from the photo, PSNR rises by 0.68\,dB and SSIM and LPIPS also improve. With a few minutes of optional fine-tuning, RAGA also surpasses per-subject methods trained on the same video. Our contributions are:
\begin{itemize}
\item A refreshable avatar formulation: appearance evidence is stored as running sums in the canonical UV input space of a decoder trained only on single images. Refreshing decodes the accumulated observations with fixed network weights, without gradient steps, image storage, or dependence on frame order.
\item A full-body system evaluated with estimation included: avatar creation in 1.08\,s and animation at 20.8\,FPS on a laptop GPU, tested on 163 clips including three held-out datasets, with a far-view protocol, a fusion-rule comparison and a multi-image baseline.
\end{itemize}

%% file: sec/2_related_work.tex
\section{Related Work} \label{sec:related} 
\paragraph{Optimization-based animatable human avatars.} Animatable avatar methods reconstruct person-specific geometry, appearance, and motion from video observations. Recent approaches build on 3D Gaussian Splatting~\cite{gaussiansplatting} for efficient real-time rendering. GART~\cite{gart} represents articulated humans using a canonical Gaussian template with learned skinning, while GaussianAvatar~\cite{gaussianavatar} jointly refines appearance and motion to capture pose-dependent effects. ExAvatar~\cite{exavatar} combines Gaussian representations with SMPL-X to support expressive body, hand, and facial animation. These methods produce high-quality personalized avatars but require video capture and person-specific optimization. In contrast, RAGA targets generalizable avatar reconstruction from a single image while retaining the ability to incorporate additional appearance observations. 

\paragraph{Generalizable human avatar reconstruction.} Generalizable methods learn reconstruction priors across identities~\cite{anigs}, allowing unseen subjects to be reconstructed without per-person training. IDOL~\cite{idol} and DiGS-Avatar~\cite{digsavatar} predict animatable Gaussian attributes in a structured UV representation, while LHM~\cite{lhm} reconstructs human Gaussians from image and body features. HumanNOVA~\cite{humannova} introduces a triplane representation conditioned on image and mesh features and trained on large-scale human data. GUAVA~\cite{guava} projects image features to template vertices and UV coordinates and decodes animatable Gaussian avatars through a feed-forward pipeline. LHM++~\cite{lhmpp} further supports pose-free reconstruction from one or multiple images. While these approaches enable accessible avatar creation, they typically produce a static reconstruction from the available observations. RAGA instead treats reconstruction as an evolving process in which newly observed appearance evidence can refine the avatar after initialization. 

\paragraph{Multi-image reconstruction.} Additional observations provide direct evidence for regions unobserved in the source image. LHM++~\cite{lhmpp} performs hierarchical point-image feature fusion across multiple images, FRESA~\cite{fresa} aggregates canonicalized observations to infer personalized geometry, skinning and deformation, and GIGA~\cite{giga} accumulates sparse-view texture evidence within a UV-aligned Gaussian representation. These methods take the image set at creation time. RAGA instead starts from one photograph and incorporates observations as they arrive.

\paragraph{Incremental and online reconstruction.} Fusing observations as they arrive has a long history in volumetric reconstruction~\cite{kinectfusion}. For head avatars, FastAvatar~\cite{fastavatar} and FFAvatar~\cite{ffavatar} reconstruct from a variable number of frames and improve as frames are added. Both aggregate the frames inside networks trained on multi-frame inputs. StreamME~\cite{streamme} and ELITE~\cite{elite} adapt head avatars with test-time optimization. Closest to our setting, GenFusion~\cite{genfusion} performs feed-forward full-body performance capture from a monocular stream: it keeps a visibility-weighted running average of features on the SMPL-X template and renders novel views of the current frame with a diffusion model trained on sequences of preceding frames. Our accumulation rule is closely related. RAGA differs in what the fused evidence produces and how its decoder is trained: the fused map regenerates an animatable Gaussian avatar that any motion can drive, the decoder is trained only on single images, and rendering runs at interactive rates.

%% file: sec/3_method.tex
\section{Method}
\label{sec:method}

\subsection{Problem Formulation and Overview}
\label{sec:overview}

\begin{figure*}[t]
    \centering
    \includegraphics[width=\textwidth]{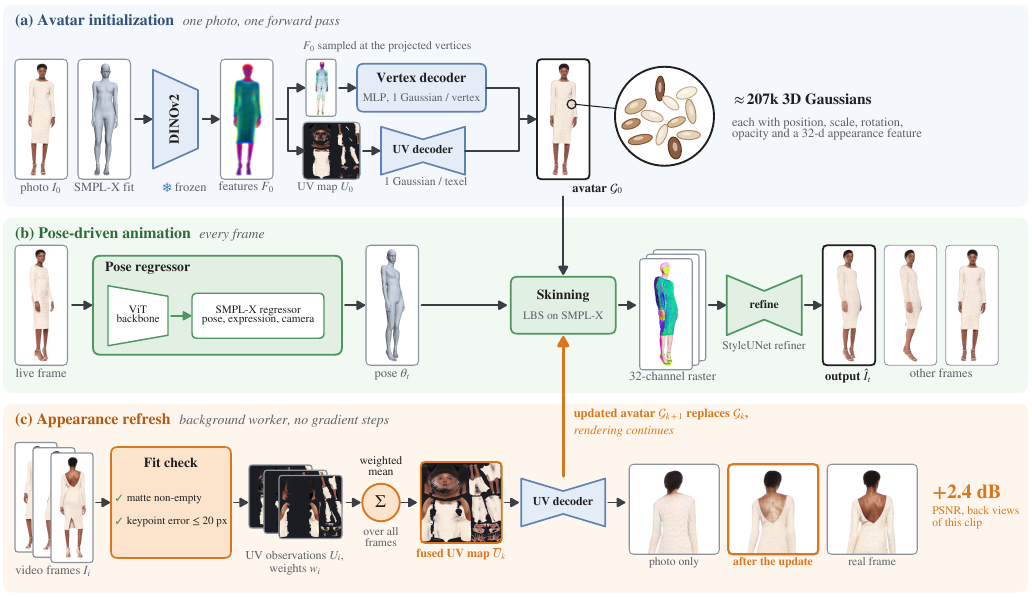}
    \vspace{-7mm}
    \caption{
    \textbf{RAGA overview.}
    \textbf{(a) Avatar initialization:}
    given one photograph and its SMPL-X fit, the reconstruction network
    initializes canonical vertex and UV Gaussians.
    \textbf{(b) Pose-driven animation:}
    estimated articulation and expression deform both Gaussian branches.
    Feature rasterization and a separate image refiner produce RGB frames.
    \textbf{(c) Appearance refresh:}
    later observations of the same person are mapped to canonical UV space
    and accumulated as weighted running sums. The unchanged UV decoder
    regenerates only the UV Gaussians. Fusion and refresh take no gradient
    steps. Body fitting for estimated parameters is iterative.
    The illustrated $+2.4$\,dB is the mean PSNR gain over five back-view
    frames of this clip, not the benchmark average.
    }
    \vspace{-4mm}
    \label{fig:raga-architecture}
\end{figure*}

Given one source image $I_0$ of a previously unseen person, RAGA
initializes an animatable avatar using a shared reconstruction prior
(Fig.~\ref{fig:raga-architecture}). Later observations supply
person-specific appearance evidence that is aligned in canonical UV
coordinates and incorporated into the avatar. Source alignment provides
session shape $\beta_0$, articulation $\theta_0$, expression $\psi_0$,
and camera $C_0$. SMPL-X~\cite{smplx} supplies a body mesh
$M(\beta,\theta,\psi)$ with shape, expression, and pose blend shapes
followed by linear blend skinning~\cite{smpl}. Later observations of the same person,
$I_i$ with $i\geq1$, have their own body alignments and cameras $C_i$.

Let $\mathcal G_k$ be the persistent canonical avatar after refresh $k$.
Rendering frame $t$ uses target articulation $\theta_t$, expression
$\psi_t$, camera $C_t$, and the latest available refresh $k(t)$:
\begin{equation}
 \widehat I_t=\mathcal S_\omega\!\left[
 \mathcal R\!\left(\mathcal W(\mathcal G_{k(t)};
 \beta_0,\theta_t,\psi_t),C_t\right)\right].
 \label{eq:raga-render}
\end{equation}
Here $\mathcal W$ deforms canonical Gaussians, $\mathcal R$ rasterizes
their features, and $\mathcal S_\omega$ refines the raster to RGB.
Animation changes posed geometry. Appearance refreshes persistently
change canonical attributes.

We describe initialization
(Sec.~\ref{sec:init}), observation fusion (Sec.~\ref{sec:appearance}),
driving (Sec.~\ref{sec:animation}), and training (Sec.~\ref{sec:training}).

\subsection{Single-Image Avatar Initialization}
\label{sec:init}

\paragraph{Backbone.}
We adopt the feed-forward architecture of GUAVA~\cite{guava}: frozen DINOv2 features, trainable projection heads, vertex and UV Gaussian decoders, and a StyleUNet~\cite{styleavatar} image refiner (\cref{eq:raga-projection,eq:raga-uvdecode}). Our changes are (i)~native SMPL-X body and head handling in place of GUAVA's inserted FLAME head~\cite{flame}, (ii)~per-image perspective projection, (iii)~full-body training (\cref{sec:training}), and (iv)~the persistent UV accumulator and refresh (\cref{sec:appearance}).
\paragraph{Canonical Gaussian Representation.} The canonical avatar consists of vertex and UV Gaussian branches,
$\mathcal G_k=\{G^V,G_k^{UV}\}$. Each Gaussian has a local displacement
$\Delta x\in\mathbb R^3$, anisotropic scale $s\in\mathbb R_+^3$, unit
quaternion $r\in\mathbb R^4$, opacity $\alpha\in[0,1]$, and appearance
feature $f\in\mathbb R^{32}$, as in Gaussian
splatting~\cite{gaussiansplatting}. Vertex centers are template
vertices: their $\Delta x=0$ is fixed, not predicted. UV Gaussians predict
local positions relative to triangle anchors.

\paragraph{Image Feature Projection.}
A frozen DINOv2 backbone~\cite{dinov2} extracts dense intermediate
features $F_i^{\mathrm{fine}}$ and a conditioning descriptor
$F_i^{\mathrm{global}}$ from the masked person image. Trainable projection
heads produce vertex features $A_i^V=H_V(F_i^{\mathrm{fine}},I_i)$ with
128 channels and UV features $A_i^{UV}=H_{UV}(F_i^{\mathrm{fine}},I_i)$
with 32 channels. For source posed vertex $\mathbf v_{0,j}$,
bilinear sampling gives $a_{0,j}=A_0^V(\pi_{C_0}(\mathbf v_{0,j}))$,
where $\pi_C$ denotes camera projection. Inverse texture mapping uses
the aligned posed mesh, triangle visibility, and barycentric coordinates:
\begin{equation}
 U_i(u)=v_i(u)[I_i\oplus A_i^{UV}]
                   (\pi_{C_i}(x_i(u)))\in\mathbb R^{35},
 \label{eq:raga-projection}
\end{equation}
% Here $u$ is a valid canonical texel, $x_i(u)$ is the posed surface point of
% texel $u$, interpolated from its triangle's vertices with barycentric
% weights, $v_i(u)\in\{0,1\}$ is visibility, and $\oplus$ is concatenation.
% The source map $U_0$ thus concatenates RGB and projected features.
% Unseen texels are zero before learned completion.
where $u$ denotes a canonical texel, $x_i(u)$ its posed surface location, $v_i(u)$ visibility, and $\oplus$ concatenation. The source map $U_0$ combines RGB and projected features, while unseen texels are initialized to zero before completion.

\paragraph{Feed-Forward Reconstruction.}
Let $B_j^V$ and $B^{UV}$ be learned vertex and UV base features. Global mappings give $z_0^V=M_V(F_0^{\mathrm{global}})$ and
$z_0^{UV}=M_{UV}(F_0^{\mathrm{global}})$, and
$e_0$ harmonically embeds the source viewing direction. The vertex decoder $D_V$ predicts $(s_j,r_j,\alpha_j,f_j)$ from
$a_{0,j}\oplus B_j^V\oplus z_0^V$ and $e_0$, and the UV decoder initializes the UV branch as

\begin{equation}
 G_0^{UV}=D_{UV}(U_0;z_0^{UV},e_0,B^{UV}).
 \label{eq:raga-uvdecode}
\end{equation}

Within $D_{UV}$, a style-conditioned StyleUNet completes the UV feature map, and a convolutional head conditioned on $e_0$ predicts Gaussian attributes at valid texels. Deformation and feature splatting of the assembled branches produce a 32-channel image. Its first three channels are directly supervised as a coarse RGB image. A separate StyleUNet image
refiner $\mathcal S_\omega$ uses all 32 channels to predict the final RGB image (Eq.~\ref{eq:raga-render}). Network dimensions and pruning are
specified in the supplementary implementation details.

\paragraph{Full-Body Adaptation.}
To support appearance alignment over the full body, we use native SMPL-X
body/head handling in place of GUAVA's inserted FLAME head, with one
consistent template for projection and animation.
For a centered square crop of side $L_i$ and focal length $f_i$ in the
same pixel units,
\begin{equation} 
(u_c,v_c) = \kappa_i(X/Z,Y/Z),\quad \kappa_i = \frac{2f_i}{L_i} = \cot\!\left(\frac{\varphi_i}{2}\right), \label{eq:raga-camera} 
\end{equation}
where $(X,Y,Z)$ are camera-space coordinates after convention conversion,
$(u_c,v_c)$ are centered normalized coordinates with $[-1,1]$ spanning
the crop, and
$\varphi_i$ is the field of view. We retain $\kappa_i=24$ when unavailable.
Each source camera governs visibility and feature projection. Each target
camera governs rasterization. Consistent body and camera handling aligns
image evidence with its UV anchors. On validation clips, the camera
change alone leaves upper-body results unchanged, while full-body
training raises PSNR on 4D-DRESS from 20.73 to 23.26\,dB, on BEDLAM2.0
from 23.65 to 25.32\,dB and on Motion-X from 23.44 to 24.58\,dB.

\subsection{Appearance Refresh}
\label{sec:appearance}
Later observations are aligned in canonical UV space and fused into a persistent appearance state, decoded with fixed network weights to refresh the avatar without retraining.
% RAGA addresses staged observation availability: a usable avatar is
% created from a single image and refined later as new observations
% become available, rather than requiring a jointly available set of
% images before initialization.
% The initial avatar is reconstructed from incomplete appearance evidence.
% Additional observations may reveal surfaces that were not visible in the
% source image, such as the back of a garment.

\paragraph{Canonical Appearance Representation.}
For observation $i$, $U_i(u)\in\mathbb{R}^{35}$ concatenates RGB and projected feature channels at canonical texel $u$ (Eq.~\ref{eq:raga-projection}). Body alignment and camera projection map all observations into a shared UV domain, where triangle identities and barycentric coordinates provide correspondence across poses. Learned completion provides a prior for regions without person-specific evidence.

\paragraph{Visibility-Aware Fusion.}
Let $\mathbf n_i(u)$ and $\mathbf d_i(u)$ be the unit triangle normal and
viewing direction, expressed in a common coordinate system. We use
\begin{equation}
 w_i(u)=v_i(u)\left|\mathbf n_i(u)^\top\mathbf d_i(u)\right|.
 \label{eq:raga-weights}
\end{equation}
The visibility term removes occluded surfaces, while the cosine term down-weights grazing views.

\paragraph{Fusing decoder inputs.}
We update observations rather than avatar parameters: UV-aligned
observations are accumulated in the decoder's input space instead of
averaging decoded Gaussians. The decoder jointly predicts appearance,
opacity and local geometry using learned priors and spatial context, so
each refresh lets its learned completion act on the combined evidence.
\Cref{sec:update-analysis} compares fusion rules.

All 35 observation channels enter the accumulator, and the source
receives weight two (results are the same with weight one):
\begin{equation} 
\begin{aligned} S_0 &= 2w_0U_0, & W_0 &= 2w_0,\\ S_n &= S_{n-1}+w_nU_n, & W_n &= W_{n-1}+w_n. \end{aligned} 
\label{eq:raga-accumulation} 
\end{equation}

\begin{equation}
\widetilde U_n(u)=
\begin{cases}
\dfrac{S_n(u)}
      {\max(W_n(u),\epsilon)},
      & W_n(u)>0,\\[4pt]
U_0(u),
      & \text{otherwise},
\end{cases}
\label{eq:raga-normalization}
\end{equation}
where $n$ counts incorporated observations and $\epsilon>0$ stabilizes
division. $S_n$ stores weighted feature sums and $W_n$ scalar weight sums,
so previous images need not be retained. New views can fill unseen texels
or revise existing evidence. The evaluated rule is a plain weighted
average with no texel-weight threshold or cap.

\paragraph{Order independence.}
Each frame's contribution $w_nU_n$ depends only on that frame, never on
other frames or on the current avatar, and a refresh reads $S_n$ and
$W_n$ without modifying them. Fusing a set of frames in one batch
therefore produces the same map, and the same refreshed avatar, as adding
the frames one at a time in any order with any number of intermediate
refreshes. Our evaluation of chronological prefixes
(\cref{sec:experiments}) thus measures the avatar that online
accumulation reaches after observing the same frames.\footnote{With
estimated parameters, each frame is fitted independently. The only
order-dependent step, reusing the previous person box when detection
fails, follows temporal order in both settings.}

\paragraph{Avatar Refresh.}
Refresh index $k$ is independent of observation index $i$ and rendering
index $t$. If $N_k$ observations have been accumulated for refresh $k+1$,
define $\overline U_k=\widetilde U_{N_k}$ and update
\begin{equation} 
\begin{aligned} G_{k+1}^{UV} &=D_{UV}(\overline U_k; z_0^{UV},e_0,B^{UV}),\\ \mathcal G_{k+1} &=\{G^V,G_{k+1}^{UV}\}. \end{aligned} 
\label{eq:raga-refresh} 
\end{equation}
Only UV Gaussians are regenerated. Canonical vertex Gaussian attributes
remain fixed. Source global conditioning, source camera conditioning,
learned bases, and all network weights remain fixed. Both branches still
deform during animation. Regeneration can change local position, scale,
rotation, and opacity as well as appearance. Unobserved texels retain
their source inputs, but spatial context in the decoder can still change
their outputs. With no additional observations, we retain $\mathcal G_0$.

With estimated parameters, each later frame is fitted iteratively with
$\beta_0$ locked and is accepted only if its matte is non-empty and its
keypoint error is at most 20\,px (Supplement).

\subsection{Pose-Driven Animation}
\label{sec:animation}
A pretrained ViTPose++-Small backbone~\cite{vitposepp} and transformer
decoder following PEAR's design~\cite{pear} predict articulation,
expression, and camera from each driving frame. Shape and camera tokens
condition the prediction, and the fitted session shape $\beta_0$ replaces
predicted shape. The regressor also initializes iterative source fitting
against detected keypoints and foreground coverage. Driving frames use
its prediction directly, without fitting.
Vertex Gaussians follow posed vertices and skinning rotations. UV
Gaussians follow their triangles' barycentric anchors, orientation, and
scale through the inherited deformation model. Equation~\ref{eq:raga-render} then
rasterizes and refines both branches. A different person's driving frames
provide motion only and are never accumulated as appearance.

\paragraph{Online operation.}
In a live session, accumulation and refresh run in a background worker while rendering continues with the latest avatar. On the laptop GPU, adding a frame takes 105\,ms for feature extraction and texture mapping and 0.8\,ms for updating the running sums. Fitting the frame with $\beta_0$ fixed takes 221\,ms, and a refresh takes 90\,ms. Measured independently without rendering, these stages total 0.42\,s. In a replay of recorded frames one update takes 1.1 to 1.3\,s, because the worker shares the GPU with the animation loop, slowing all stages, and performs extra validation by fitting some frames and re-rendering the avatar to compare against incoming observations. During the same replay, the application runs at 23.2\,FPS with the worker disabled and 13.4\,FPS with it enabled. It estimates the pose of the next frame in a second thread while rendering the current one, so its rate with the worker disabled exceeds the 20.8\,FPS given by the sequential 48.1\,ms per frame of \cref{tab:runtime}. These validation checks, along with the worker's frame and texel gating heuristics, are not used in any reported result.

\begin{table*}[t]
\centering
\definecolor{ragaCellBest}{RGB}{248,205,205}
\definecolor{ragaCellSecond}{RGB}{205,220,245}
\caption{\textbf{Reconstruction quality per dataset and on all 163 clips.} Clip counts are in parentheses. All rows use supplied driving parameters. The first block initializes from one photo without training on the test person. The refresh (photo + update) also uses later images of that person with fixed network weights. Because fusion is order-independent, fusing the reconstruction frames in one batch gives the avatar that online accumulation of the same frames reaches. LHM++ with 16 images receives the source photo and 15 evenly spaced reconstruction frames. $^\dagger$Held-out datasets, never used in training. GUAVA is trained on upper bodies only. Video methods train on the reconstruction partition. Metrics average frames within each clip and then average clips equally. ExAvatar also receives supplied facial parameters. \colorbox{ragaCellBest}{\textbf{Best}} and \colorbox{ragaCellSecond}{second best} cells are shaded within each block, and ties share a rank.}
\vspace{-2mm}
\label{tab:main}
\setlength{\tabcolsep}{3.5pt}
\renewcommand{\arraystretch}{1.15}
\resizebox{\textwidth}{!}{%
\begin{tabular}{|l|l|ccc|ccc|ccc|ccc|ccc|ccc|}
\hline
\multicolumn{1}{|l|}{} & \multicolumn{1}{l|}{} & \multicolumn{3}{c|}{UBody (30)} & \multicolumn{3}{c|}{People-Snapshot$^\dagger$ (24)} & \multicolumn{3}{c|}{NeuMan$^\dagger$ (6)} & \multicolumn{3}{c|}{X-Humans$^\dagger$ (3)} & \multicolumn{3}{c|}{UBC-Fashion (100)} & \multicolumn{3}{c|}{\textbf{All clips (163)}} \\
\cline{3-20}
Method & Appearance input & {\scriptsize PSNR$\uparrow$} & {\scriptsize SSIM$\uparrow$} & {\scriptsize LPIPS$\downarrow$} & {\scriptsize PSNR$\uparrow$} & {\scriptsize SSIM$\uparrow$} & {\scriptsize LPIPS$\downarrow$} & {\scriptsize PSNR$\uparrow$} & {\scriptsize SSIM$\uparrow$} & {\scriptsize LPIPS$\downarrow$} & {\scriptsize PSNR$\uparrow$} & {\scriptsize SSIM$\uparrow$} & {\scriptsize LPIPS$\downarrow$} & {\scriptsize PSNR$\uparrow$} & {\scriptsize SSIM$\uparrow$} & {\scriptsize LPIPS$\downarrow$} & {\scriptsize PSNR$\uparrow$} & {\scriptsize SSIM$\uparrow$} & {\scriptsize LPIPS$\downarrow$} \\
\hline\hline
\multicolumn{20}{|l|}{\textit{Feed forward, no training on the test person}} \\
\hline
GUAVA~\cite{guava} & photo & 19.48 & \cellcolor{ragaCellSecond}0.788 & \cellcolor{ragaCellSecond}0.191 & 25.58 & 0.917 & 0.079 & 23.36 & 0.899 & \cellcolor{ragaCellSecond}0.081 & 22.75 & 0.906 & 0.091 & 22.77 & 0.918 & 0.058 & 22.60 & 0.893 & 0.087 \\
\hline
LHM~\cite{lhm} & photo & 16.12 & 0.692 & 0.263 & 25.11 & 0.914 & 0.085 & 22.43 & 0.889 & 0.090 & 24.11 & 0.913 & 0.091 & 22.13 & 0.911 & 0.063 & 21.51 & 0.870 & 0.105 \\
\hline
LHM++~\cite{lhmpp} & photo & 16.72 & 0.724 & 0.240 & 24.68 & 0.914 & 0.086 & 22.32 & 0.892 & 0.088 & 22.93 & 0.913 & 0.083 & 21.21 & 0.909 & 0.065 & 20.97 & 0.875 & 0.102 \\
\hline
LHM++~\cite{lhmpp} & 16 photos & 16.94 & 0.735 & 0.235 & 24.79 & 0.915 & 0.082 & 22.50 & 0.894 & 0.087 & 23.47 & 0.916 & 0.078 & 21.34 & 0.910 & 0.063 & 21.12 & 0.878 & 0.099 \\
\hline
IDOL~\cite{idol} & photo & 14.74 & 0.636 & 0.308 & 24.38 & 0.910 & 0.085 & 22.27 & 0.886 & 0.095 & 23.76 & 0.915 & 0.084 & 21.80 & 0.910 & 0.066 & 20.93 & 0.859 & 0.114 \\
\hline
RAGA & photo & \cellcolor{ragaCellSecond}19.62 & 0.786 & \cellcolor{ragaCellSecond}0.191 & \cellcolor{ragaCellSecond}26.54 & \cellcolor{ragaCellSecond}0.924 & \cellcolor{ragaCellBest}\textbf{0.071} & \cellcolor{ragaCellSecond}23.77 & \cellcolor{ragaCellSecond}0.901 & \cellcolor{ragaCellBest}\textbf{0.076} & \cellcolor{ragaCellSecond}24.90 & \cellcolor{ragaCellSecond}0.922 & \cellcolor{ragaCellBest}\textbf{0.075} & \cellcolor{ragaCellSecond}24.35 & \cellcolor{ragaCellSecond}0.930 & \cellcolor{ragaCellBest}\textbf{0.046} & \cellcolor{ragaCellSecond}23.79 & \cellcolor{ragaCellSecond}0.902 & \cellcolor{ragaCellBest}\textbf{0.078} \\
\hline
RAGA & photo + update & \cellcolor{ragaCellBest}\textbf{20.74} & \cellcolor{ragaCellBest}\textbf{0.810} & \cellcolor{ragaCellBest}\textbf{0.181} & \cellcolor{ragaCellBest}\textbf{27.01} & \cellcolor{ragaCellBest}\textbf{0.931} & \cellcolor{ragaCellSecond}0.075 & \cellcolor{ragaCellBest}\textbf{24.43} & \cellcolor{ragaCellBest}\textbf{0.907} & 0.083 & \cellcolor{ragaCellBest}\textbf{25.26} & \cellcolor{ragaCellBest}\textbf{0.928} & \cellcolor{ragaCellSecond}0.077 & \cellcolor{ragaCellBest}\textbf{24.65} & \cellcolor{ragaCellBest}\textbf{0.932} & \cellcolor{ragaCellSecond}0.049 & \cellcolor{ragaCellBest}\textbf{24.28} & \cellcolor{ragaCellBest}\textbf{0.908} & \cellcolor{ragaCellSecond}0.079 \\
\hline
\hline
\multicolumn{20}{|l|}{\textit{Training on the test person}} \\
\hline
GART~\cite{gart} & video & \cellcolor{ragaCellSecond}19.57 & \cellcolor{ragaCellSecond}0.797 & \cellcolor{ragaCellSecond}0.186 & \cellcolor{ragaCellSecond}29.57 & \cellcolor{ragaCellBest}\textbf{0.948} & 0.052 & \cellcolor{ragaCellSecond}25.46 & \cellcolor{ragaCellSecond}0.917 & \cellcolor{ragaCellSecond}0.073 & \cellcolor{ragaCellSecond}26.43 & \cellcolor{ragaCellBest}\textbf{0.935} & \cellcolor{ragaCellSecond}0.071 & \cellcolor{ragaCellSecond}23.90 & \cellcolor{ragaCellSecond}0.927 & \cellcolor{ragaCellSecond}0.052 & \cellcolor{ragaCellSecond}24.04 & \cellcolor{ragaCellSecond}0.906 & \cellcolor{ragaCellSecond}0.078 \\
\hline
GaussianAvatar~\cite{gaussianavatar} & video & 19.02 & 0.785 & 0.204 & 27.40 & 0.932 & 0.069 & 23.91 & 0.903 & 0.083 & 25.05 & \cellcolor{ragaCellSecond}0.923 & \cellcolor{ragaCellSecond}0.071 & 23.17 & 0.922 & 0.062 & 23.09 & 0.898 & 0.090 \\
\hline
ExAvatar~\cite{exavatar} & video & 19.47 & \cellcolor{ragaCellSecond}0.797 & 0.199 & 28.58 & 0.942 & \cellcolor{ragaCellSecond}0.051 & 24.81 & 0.913 & 0.076 & 23.18 & 0.911 & 0.087 & 23.52 & 0.925 & 0.057 & 23.56 & 0.903 & 0.084 \\
\hline
RAGA, fine tuned & video & \cellcolor{ragaCellBest}\textbf{22.69} & \cellcolor{ragaCellBest}\textbf{0.839} & \cellcolor{ragaCellBest}\textbf{0.140} & \cellcolor{ragaCellBest}\textbf{30.12} & \cellcolor{ragaCellSecond}0.945 & \cellcolor{ragaCellBest}\textbf{0.048} & \cellcolor{ragaCellBest}\textbf{26.32} & \cellcolor{ragaCellBest}\textbf{0.919} & \cellcolor{ragaCellBest}\textbf{0.060} & \cellcolor{ragaCellBest}\textbf{26.86} & \cellcolor{ragaCellBest}\textbf{0.935} & \cellcolor{ragaCellBest}\textbf{0.055} & \cellcolor{ragaCellBest}\textbf{24.97} & \cellcolor{ragaCellBest}\textbf{0.935} & \cellcolor{ragaCellBest}\textbf{0.041} & \cellcolor{ragaCellBest}\textbf{25.39} & \cellcolor{ragaCellBest}\textbf{0.918} & \cellcolor{ragaCellBest}\textbf{0.061} \\
\hline
\end{tabular}}
\vspace{-4mm}
\end{table*}

\subsection{Training}
\label{sec:training}
\paragraph{Training data.}
Starting from the released GUAVA weights~\cite{guava}, we train on
BEDLAM2.0~\cite{bedlam2} (1{,}500 synthetic clips, about a quarter of
training samples), UBody~\cite{ubody} (203 clips),
Motion-X~\cite{motionx} (310 clips), 4D-DRESS~\cite{fourddress}
(28 subjects) and 110 of the 500 UBC-Fashion~\cite{ubcfashion} training
videos. To keep test people unseen, we drop the 6 videos that
GART~\cite{gart} evaluates on and 362 that share a person with them or
with a test video. Another 22 fail a body alignment check. No UBody
test identity appears in training, and People-Snapshot~\cite{peoplesnapshot}, NeuMan~\cite{neuman} and
X-Humans~\cite{xhumans} are held out entirely. Training always builds the avatar from
a single image, so the decoder never sees fused maps (Supplement).

\paragraph{Losses.}
We train reconstruction on same-person source-target pairs. DINOv2 remains frozen. Projection heads,
global mappings, learned bases, Gaussian decoders, and the image refiner
are optimized using
\begin{equation}
 \begin{split}
 \mathcal L={}&\mathcal L_{\mathrm{rec}}+\mathcal L_{\mathrm{reg}}
       +0.05\mathcal L_{\mathrm{id}}\\
       &+0.5\mathcal L_{\mathrm{cloth}}
       +0.5\mathcal L_{\mathrm{edge}}+0.5\mathcal L_{\mathrm{sil}}.
 \end{split}
 \label{eq:raga-loss}
\end{equation}
The reconstruction term supervises raw and refined RGB with
$\ell_1+0.05\,\mathrm{LPIPS}$~\cite{lpips}, including head and hand crops
with multipliers 0.5 and 0.1. The regularization term penalizes excessive
UV local displacement and scale. Identity supervision uses ArcFace
cosine distance~\cite{arcface} on sufficiently large, approximately
frontal faces. Clothing supervision applies the reconstruction metric
within the clothing region. Edge and silhouette terms use $\ell_1$
penalties on aligned self-target Sobel responses and foreground coverage,
respectively.

The pose regressor is trained separately with parameter, 3D-joint, and
2D-keypoint supervision. Optional per-person fine-tuning updates
reconstruction weights using reconstruction, regularization, and identity
losses (Supplement). Observation fusion keeps network weights fixed.

%% file: sec/4_result.tex
\definecolor{ragaBest}{RGB}{180,35,35}
\definecolor{ragaSecond}{RGB}{35,80,170}
\definecolor{ragaThird}{RGB}{20,115,65}
\providecommand{\ragaFirst}[1]{\textcolor{ragaBest}{\textbf{#1}}}
\providecommand{\ragaSecond}[1]{\textcolor{ragaSecond}{#1}}
\providecommand{\ragaThird}[1]{\textcolor{ragaThird}{#1}}

\section{Results}
\label{sec:experiments}

\paragraph{Evaluation Setup.}
\label{sec:experimental-setup}
We evaluate on 163 clips comprising 30 from UBody, 24 from
People-Snapshot, 6 from NeuMan, 3 from X-Humans and 100 from UBC-Fashion. Each dataset has separate reconstruction and test partitions, with the
first reconstruction frame used as the source image. All methods are
scored on the same $512\times512$ person crops against masked targets
on black. We average PSNR, SSIM and LPIPS-VGG within each clip and then
across clips. Paired 95\% confidence intervals use 10,000 bootstrap
samples of people within each dataset.
All methods are driven by the same supplied body parameters, converted to each method's body model, and all baselines are rerun with their released models under this protocol (Supplement). For UBC-Fashion, the supplied parameters were fitted using the whole video, including test frames, and shared across methods. Appearance training and fusion use only reconstruction frames.

% The main comparison uses common supplied driving parameters, converted
% to each method's body model. These include fits and pseudo labels.
% All baselines are rerun with their released models under this protocol,
% without pose or expression optimization on test frames, so their scores
% differ from the published ones.
% For UBC-Fashion, the supplied parameters were fitted using the whole
% video, including test frames, and shared across methods. Appearance
% training and fusion use only reconstruction frames. We distinguish
% single image reconstruction, updates with fixed network weights and
% training on each test person.

\subsection{Quantitative Results}
\label{sec:comparisons}

\begin{table}[t]
\centering
\caption{\textbf{Views the photo never showed.} The 1{,}576 test frames
turned at least $135^\circ$ from the source (130 clips, UBody has no
such views), on UBC-Fashion, on the three held-out datasets and on all
clips. Clip counts are in parentheses. All rows use supplied driving
parameters. Bold marks the best score in each block.}
\vspace{-2mm}
\label{tab:farview}
\setlength{\tabcolsep}{3pt}
\resizebox{\columnwidth}{!}{%
\begin{tabular}{@{}lccccccccc@{}}
\toprule
 & \multicolumn{3}{c}{UBC-Fashion (99)} & \multicolumn{3}{c}{Held out (31)} & \multicolumn{3}{c}{All (130)} \\
\cmidrule(lr){2-4}\cmidrule(lr){5-7}\cmidrule(l){8-10}
Method & {\scriptsize PSNR$\uparrow$} & {\scriptsize SSIM$\uparrow$} & {\scriptsize LPIPS$\downarrow$} & {\scriptsize PSNR$\uparrow$} & {\scriptsize SSIM$\uparrow$} & {\scriptsize LPIPS$\downarrow$} & {\scriptsize PSNR$\uparrow$} & {\scriptsize SSIM$\uparrow$} & {\scriptsize LPIPS$\downarrow$} \\
\midrule
\multicolumn{10}{@{}l}{\textit{Feed forward, no training on the test person}} \\
GUAVA~\cite{guava} & 20.63 & 0.900 & 0.083 & 23.94 & 0.903 & 0.102 & 21.42 & 0.901 & 0.088 \\
LHM~\cite{lhm} & 21.35 & 0.904 & 0.073 & 24.32 & 0.906 & 0.096 & 22.06 & 0.904 & 0.079 \\
LHM++~\cite{lhmpp} & 20.37 & 0.903 & 0.075 & 23.87 & 0.907 & 0.097 & 21.21 & 0.904 & 0.081 \\
IDOL~\cite{idol} & 20.83 & 0.903 & 0.076 & 23.60 & 0.903 & 0.097 & 21.49 & 0.903 & 0.081 \\
RAGA & 22.30 & 0.916 & 0.064 & 25.11 & 0.911 & 0.091 & 22.97 & 0.914 & 0.071 \\
RAGA + update & \textbf{22.90} & \textbf{0.919} & \textbf{0.060} & \textbf{26.04} & \textbf{0.921} & \textbf{0.087} & \textbf{23.65} & \textbf{0.919} & \textbf{0.067} \\
\midrule
\multicolumn{10}{@{}l}{\textit{Training on the test person}} \\
GART~\cite{gart} & 21.60 & 0.904 & 0.075 & 27.98 & \textbf{0.935} & 0.070 & 23.12 & 0.911 & 0.074 \\
GaussianAvatar~\cite{gaussianavatar} & 20.98 & 0.903 & 0.080 & 26.24 & 0.921 & 0.081 & 22.23 & 0.907 & 0.080 \\
ExAvatar~\cite{exavatar} & 21.33 & 0.905 & 0.074 & 26.99 & 0.928 & 0.066 & 22.68 & 0.911 & 0.072 \\
RAGA, fine tuned & \textbf{23.15} & \textbf{0.919} & \textbf{0.057} & \textbf{28.67} & 0.934 & \textbf{0.061} & \textbf{24.47} & \textbf{0.922} & \textbf{0.058} \\
\bottomrule
\end{tabular}}
\vspace{-4mm}
\end{table}

\begin{figure*}[t]
\centering
\includegraphics[width=0.9\textwidth]{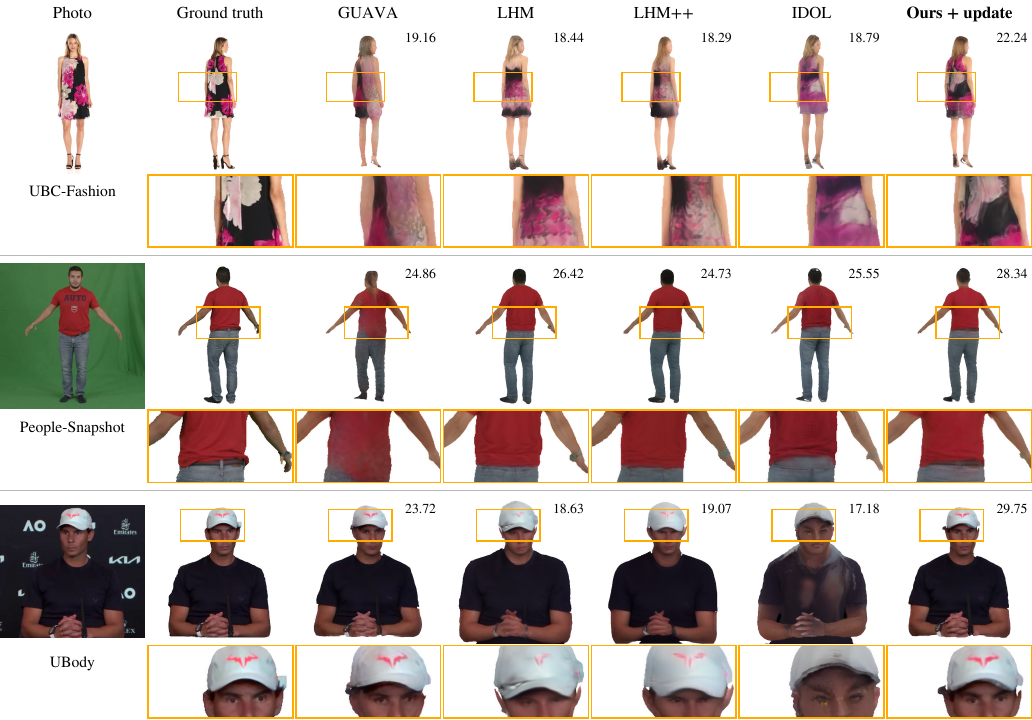}
\vspace{-2mm}
\caption{\textbf{Selected qualitative comparisons.}
GUAVA, LHM, LHM++ and IDOL use one source image. RAGA with update also
uses later observations of the same person with fixed network weights.
All methods use supplied driving parameters. Values are PSNR for the
illustrated target frames. 
Enlarged regions show garment appearance in detail.}
\vspace{-5mm}
\label{fig:qualitative}
\end{figure*}

\paragraph{Single photo.}
From one image, RAGA reaches 23.79\,dB PSNR, 0.902 SSIM and
0.078 LPIPS, the best among photo-only methods (\cref{tab:main}), and
has the lowest LPIPS on four of five datasets. Its advantage does not depend on overlap with the training domain: on the 33 clips of the three held-out datasets, it reaches 25.89\,dB against 24.53, 24.09 and
23.94\,dB for LHM, LHM++ and IDOL, a lead of 1.36\,dB $[0.99,1.71]$,
1.80\,dB $[1.43,2.16]$ and 1.94\,dB $[1.65,2.23]$ (Supplement). GUAVA is trained on upper bodies only, so full-body clips lie outside its training distribution. We include it as a backbone reference, not a full-body competitor. 
% RAGA improves on it by 1.19\,dB $[1.06,1.32]$.

\paragraph{Refresh.}
Refreshing from later frames, with fixed network weights, raises RAGA to
24.28\,dB PSNR and 0.908 SSIM. PSNR and SSIM improve on every dataset.
LPIPS improves on UBody and rises slightly elsewhere, from 0.078 to 0.079
overall, but improves on views the photo never showed (\cref{tab:farview}). 
LHM++ can also take several images. Given 16 images per clip (the source
photo and 15 evenly spaced reconstruction frames) on all 163 clips, its
PSNR rises from 20.97 to 21.12\,dB (\cref{tab:main}). Fusing the same 16
images raises RAGA from 23.79 to 24.27\,dB, which is 3.15\,dB
$[2.88,3.43]$ above LHM++ with 16 images and within 0.01\,dB of fusing
every reconstruction frame (Supplement).

\paragraph{Gains concentrate on unseen surfaces.}
Most test frames show surfaces the source photo already covers, so the
all-frame average understates the effect on newly observed regions.
\Cref{tab:farview} evaluates the 1{,}576 test frames turned at least
$135^\circ$ from the source, covering 130 clips (UBody has no such
views). RAGA from one photo is already the best single-image method on
these views, and refreshing improves all three metrics, including LPIPS.
The PSNR gain is positive on every dataset and largest on the held-out
ones: 0.97\,dB on People-Snapshot, 0.86\,dB on NeuMan, 0.76\,dB on
X-Humans and 0.60\,dB on UBC-Fashion (Supplement). Over these clips, UV coverage
rises from 41\% to 85\%.  Since 99 of the 130 clips are from UBC-Fashion, \cref{tab:farview} also reports them apart from the 31 held-out clips. On UBC-Fashion the refreshed avatar is 1.29\,dB $[0.99,1.59]$ above GART. On held-out clips GART stays ahead by 1.94\,dB $[1.54,2.24]$, and only fine-tuned RAGA passes it, by 0.69\,dB $[0.40,0.93]$.

\paragraph{Per-subject methods.} The refreshed avatar exceeds GART overall in PSNR (24.28 versus
24.04\,dB) and SSIM (0.908 versus 0.906). This advantage comes from the in-domain datasets: on the held-out datasets, per-subject optimization
remains ahead (26.39 versus 28.54\,dB for GART, a gap of 2.15\,dB $[1.78,2.42]$). With 500 optimization steps per subject, RAGA reaches 25.39\,dB PSNR, 0.918 SSIM and 0.061 LPIPS, outperforming GART by 1.35\,dB $[1.10,1.67]$, and also leads on the held-out datasets (29.13 versus 28.54\,dB, a lead of 0.59\,dB $[0.46,0.73]$).
% Fine-tuning takes a median of 4.0 minutes per clip on a shared L40S GPU, where all training was run, against 2.3 minutes for GART, 47 for GaussianAvatar and 76 for ExAvatar.

\paragraph{Runtime.}
\label{sec:efficiency}
\Cref{tab:runtime} reports inference timings on an NVIDIA GeForce RTX 3080 Laptop
GPU. RAGA's 1.08\,s creation time includes person detection (0.04\,s),
the source fit (0.81\,s: keypoints, matting, the pose regressor and 150
fitting steps), preprocessing (0.03\,s) and the avatar network
(0.20\,s). Its 48.1\,ms per frame cover cropping with a fixed centre
box, the pose regressor (13.7\,ms), skinning, rasterization and the refiner. With a person
detector on every frame, animation takes 85.4\,ms. Of GUAVA's 7.28\,s per frame, its tracker takes
7.25\,s. Times exclude model loading, capture, image decoding and
display for all methods (Supplement).

\begin{table}[t]
\centering
\caption{\textbf{Runtime on an NVIDIA GeForce RTX 3080 Laptop GPU.} Reconstruction includes source-image processing, and animation reports per-frame time. Source pose and matte indicate whether inputs are estimated or supplied. Supplied inputs are not timed. For RAGA and GUAVA, animation includes per-frame pose estimation. LHM~1B exceeded GPU memory during reconstruction and could not be animated. LHM++ uses flash attention to fit in memory. Best result is highlighted.}
\vspace{-2mm}
\label{tab:runtime}
\setlength{\tabcolsep}{4pt}
\renewcommand{\arraystretch}{1.1}
\resizebox{\columnwidth}{!}{%
\begin{tabular}{@{}lllrr@{}}
\toprule
Method & Source pose & Matte & Rec.\ (s) & Anim.\ (ms) \\
\midrule
LHM 1B & supplied & supplied & OOM & N/R \\
LHM++ 700M & supplied & supplied & 3.99 & 170.4 \\
IDOL & not needed & supplied & 8.87 & 69.0 \\
GUAVA + EHM tracker & estimated & computed & 11.11 & 7{,}280 \\
RAGA + PEAR & estimated & computed & 1.12 & 94.0 \\
RAGA + our pose regressor & estimated & computed & \textbf{1.08} & \textbf{48.1} \\
\bottomrule
\end{tabular}}
\vspace{-4mm}
\end{table}

\paragraph{Pose regressor.}
The network alone runs at 154\,FPS on this GPU, against 18.4\,FPS for PEAR. Its 2D accuracy on COCO and LSP-Extended matches or exceeds OSX,
Multi-HMR~\cite{multihmr} and SMPLest-X~\cite{smplestx} but is below PEAR, and its 3DPW errors are higher
than those of all compared methods. The supplement gives full results.

\subsection{Qualitative Results}
\label{sec:qualitative-results}

\Cref{fig:qualitative} shows one example from each of three datasets. On
UBC-Fashion, the updated avatar reproduces the large floral pattern on
the back of the dress more accurately, although fine texture remains
imperfect. On People-Snapshot, it recovers the belt seen from behind.
The UBody example is a front view: LHM, LHM++ and IDOL blur the cap
logo, and GUAVA renders it slightly tilted with one eye blurred, while
the updated avatar keeps the logo in place and the eyes sharp. The
paired RAGA results in Table~\ref{tab:main} measure the effect of
appearance updating within our model.
\begin{table}[t]
\centering
\caption{\textbf{Ablation of appearance updating.} Frames used denotes the fraction of reconstruction frames fused after the source photo. Scores are averaged over 163 clips. $\Delta$PSNR and 95\% confidence intervals are measured relative to no update (0\%) within each block. Bold indicates the best score.}
\vspace{-2mm}
\label{tab:ablation}
\small
\setlength{\tabcolsep}{2pt}
\begin{tabular*}{\columnwidth}
{@{\extracolsep{\fill}}lcccc@{}}
\toprule
\begin{tabular}[c]{@{}l@{}}
Frames\\used
\end{tabular}
& PSNR$\uparrow$
& SSIM$\uparrow$
& LPIPS$\downarrow$
& $\Delta$PSNR [95\% CI] \\
\midrule
\multicolumn{5}{@{}l}{
\textit{Supplied body and camera parameters}} \\
0\%   & 23.79 & 0.902 & \textbf{0.078} & reference \\
25\%  & 24.08 & 0.906 & \textbf{0.078}
      & $+0.29\,[0.23,\,0.36]$ \\
50\%  & 24.20 & 0.907 & 0.079
      & $+0.41\,[0.33,\,0.49]$ \\
100\% & \textbf{24.28} & \textbf{0.908} & 0.079
      & $+0.49\,[0.40,\,0.59]$ \\
\midrule
\multicolumn{5}{@{}l}{
\textit{Estimated body and camera parameters}} \\
0\%   & 19.89 & 0.865 & \textbf{0.109} & reference \\
100\% & \textbf{20.26} & \textbf{0.870} & 0.111
      & $+0.37\,[0.31,\,0.42]$ \\
\midrule
\multicolumn{5}{@{}l}{
\textit{Fine tuned with supplied parameters}} \\
0\%   & \textbf{25.39} & \textbf{0.918}
      & \textbf{0.061} & reference \\
100\% & 24.84 & 0.915 & 0.070
      & $-0.55\,[-0.70,\,-0.41]$ \\
\bottomrule
\end{tabular*}
\vspace{-4mm}
\end{table}

\subsection{Analysis of the Refresh}
\label{sec:update-analysis}

\paragraph{Amount of evidence.}
Fusing 25\%, 50\% and 100\% of the reconstruction frames raises UV
coverage over all 163 clips from 40\% for the photo to 53\%, 59\% and
78\%, with PSNR gains of 0.29, 0.41 and 0.49\,dB (\cref{tab:ablation}; Supplement).
The smaller gains suggest diminishing returns, although these
prefixes change both frame count and viewpoint coverage.
\noindent
\paragraph{Fusion rule.}
Both kinds of evidence contribute. On a separate validation set of 20
UBody and 30 BEDLAM2.0 clips, filling only texels unseen in the photo
raises PSNR from 20.98 to 21.09\,dB, and letting new views also revise
texels the photo covered raises it to 21.59\,dB. Keeping each texel's
best facing view (21.02\,dB) or adding a texel-weight threshold and cap
(21.20\,dB) does worse than the plain weighted mean (Supplement). The UBody
clips in this set never turn more than $39^\circ$ from the photo, which
limits how much unseen surface they contain.
\noindent
\paragraph{Estimated parameters.}
When body and camera parameters are estimated from the images, our
pipeline fits the source image and the later frames, and the pose
regressor drives each test frame. Refreshing raises PSNR from 19.89 to
20.26\,dB, a gain of 0.37\,dB $[0.31,0.42]$, while LPIPS changes from
0.109 to 0.111. The result remains well below the supplied-parameter
setting. This comparison changes initialization, alignment and driving
together.

\paragraph{After fine-tuning.}
Fusing after per-subject fine-tuning lowers PSNR from 25.39 to 24.84\,dB,
a paired change of $-0.55$\,dB $[-0.70,-0.41]$, and raises LPIPS from
0.061 to 0.070. The two do not combine under this protocol.

%% file: sec/5_conclusion.tex
\paragraph{Limitations.}
Refreshing slightly raises LPIPS on views the photo already covered. In a
diagnosed case, texels seen edge-on at the silhouette pick up background
colour through the matte boundary. With estimated body parameters,
quality remains well below the supplied-parameter setting, and baselines
were not evaluated in that setting. Fusion after fine-tuning lowers
quality, and the live worker's gating heuristics are not validated.

\section{Conclusion}
\label{sec:conclusion}
RAGA shows that a single-photo avatar need not be a one-time prediction.
By accumulating later observations in the input space of a decoder
trained only on single images, it refreshes a full-body avatar without
retraining, recovers appearance the photo never showed, and runs creation
and animation on a laptop GPU with body and pose estimation included.

%% file: sec/llm_usage.tex
\section*{LLM Usage Statement}
OpenAI Codex and Claude (Anthropic) assisted with literature search and review, evaluation of the proposed method in the context of existing research, preliminary experimental analysis, code development and debugging, and drafting, editing, and revision of the manuscript and supplementary material. The authors retain full responsibility for the research conception, methodology, experiments, scientific claims, results, references, and final content of the paper.